\documentclass[letterpaper, 10 pt, conference]{ieeeconf}  

\IEEEoverridecommandlockouts                              

\usepackage{graphics} 
\usepackage{epsfig} 
\usepackage{mathptmx} 
\usepackage{times} 
\usepackage{amsmath} 
\usepackage{amssymb}  
\usepackage{hyperref}
\usepackage{cleveref}
\usepackage{siunitx}
\usepackage{booktabs}
\usepackage{threeparttable}
\usepackage[table]{xcolor}
\usepackage{wrapfig}

\title{
{\normalsize \texttt{[This is a preprint - Accepted for publication at IROS 2025]}}\\
{\LARGE \bf Towards Data-Driven Adaptive Exoskeleton Assistance\\for Post-stroke Gait}
}

\author{Fabian C. Weigend$^{*}$, Dabin K. Choe$^{*}$, Santiago Canete$^{*}$, Conor J. Walsh$^{\dagger}$
\thanks{This work was supported by the Harvard University John A. Paulson School of Engineering and Applied Sciences.}
\thanks{This work involved human subjects or animals in its research. Approval
of all ethical and experimental procedures and protocols was granted by
Harvard Medical School Institutional Review Board.}%
\thanks{All authors are affiliated with the Harvard John A. Paulson School of Engineering and Applied Sciences,
Boston, MA, USA.}
\thanks{$*$: F. C. Weigend, D. K. Choe, S. Canete contributed equally to this work}%
\thanks{$\dagger$: Corresponding author: walsh@seas.harvard.edu}%
}

\begin{document}

\maketitle

\thispagestyle{empty}
\pagestyle{empty}

\begin{abstract}
Recent work has shown that exoskeletons controlled through data-driven methods can dynamically adapt assistance to various tasks for healthy young adults. However, applying these methods to populations with neuromotor gait deficits, such as post-stroke hemiparesis, is challenging. This is due not only to high population heterogeneity and gait variability but also to a lack of post-stroke gait datasets to train accurate models. Despite these challenges, data-driven methods offer a promising avenue for control, potentially allowing exoskeletons to function safely and effectively in unstructured community settings. This work presents a first step towards enabling adaptive plantarflexion and dorsiflexion assistance from data-driven torque estimation during post-stroke walking. We trained a multi-task Temporal Convolutional Network (TCN) using collected data from four post-stroke participants walking on a treadmill ($\mathrm{R}^2$ of 0.74 ± 0.13). The model uses data from three inertial measurement units (IMU) and was pretrained on healthy walking data from 6 participants. We implemented a wearable prototype for our ankle torque estimation approach for exoskeleton control and demonstrated the viability of real-time sensing, estimation, and actuation with one post-stroke participant.
\end{abstract}

\section{INTRODUCTION}

Stroke is the leading cause of gait impairment, significantly reducing mobility, independence, and overall quality of life \cite{mulroy_use_2003,kuch_identification_2025}.
After a stroke, mobility recovery tends to plateau within three to five months, leaving many individuals reliant on ankle-foot-orthoses (AFOs) to ambulate safely \cite{leung_impact_2003}.
Although AFOs provide medial-lateral support and prevent foot-drop, their rigid structure hinders the user's ability to plantarflex the ankle \cite{mulroy_effect_2010}, and propel their body forward.
Consequently, research into active devices that can both support the ankle and augment push-off could lead to assistive devices that allow for safer, faster, and more efficient ambulation.

Exoskeletons are promising tools for improving gait function after a stroke. By assisting the paretic ankle, stroke survivors can increase foot clearance \cite{sloot_effects_2023} and forward propulsion \cite{awad_soft_2017}, resulting in faster \cite{awad_walking_2020} and more efficient \cite{awad_reducing_2017} ambulation. However, current methods for controlling these devices are limited in their capacity to adapt to the user, the environment, and the task. A common control strategy relies on a feed-forward approach in which a pre-determined assistive profile suitable for the specified task is delivered at certain gait events \cite{mooney_autonomous_2014,zhang_human_2017,bae_lightweight_2018}. Once determined, this profile remains fixed, independently of the user's movement pattern. Although this approach has been effective in constrained settings it is not suitable for use in less-structured environments, especially for populations with high gait variability. For this reason, it is necessary to explore adaptive control methods that can capture the variability in post-stroke gait and could allow for translation outside the laboratory.



In healthy populations, wearable exoskeletons have made progress towards more adaptive solutions. 
Recently, data-driven approaches for estimating knee and hip torques with wearable sensors were used to adapt exoskeleton assistance for a variety of tasks in healthy young adults \cite{molinaro_estimating_2024,molinaro_task-agnostic_2024}.
With these data-driven methods, assistance can be personalized to the user under real-world conditions more rapidly than using conventional approaches \cite{slade_personalizing_2022,slade_rapid_2019}, demonstrating the potential of data-driven methods for capturing the variability in human movement.
However, the applicability of these methods for exoskeleton control is yet to be validated for clinical populations, which may be challenging due to high gait variability both within and across individuals.

\begin{figure*}[!htbp]
      \centering
      \includegraphics[width=\linewidth]{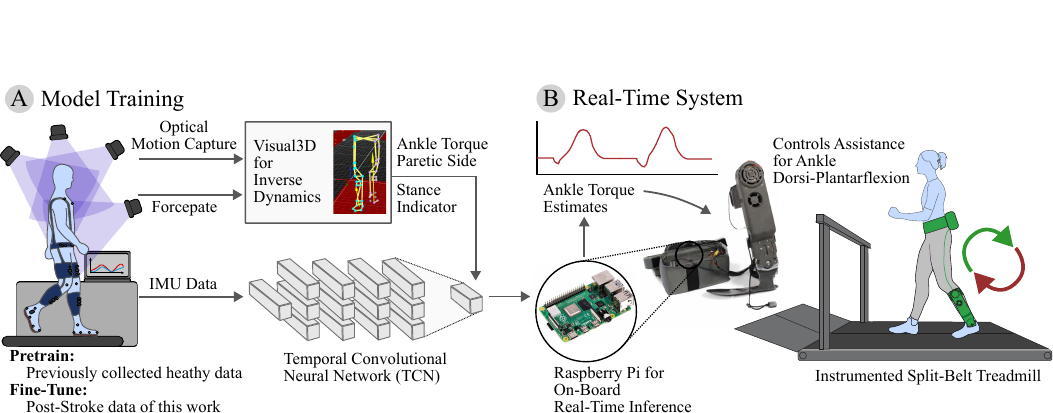}
      \caption{\textbf{A:} We trained a Temporal Convolutional Neural Network (TCN) to estimate the ankle moment of the paretic leg from IMU sensor data. The model was trained using previously collected treadmill walking data from healthy and post-stroke participants. \textbf{B:} We deployed the model on a microcomputer for real-time estimates to enable adaptive assistance in dorsi-plantarflexion for an ankle exoskeleton. The controller used a scaling gain and directly commanded the torque to the device in closed-loop.}
      \label{fig:methods_overview}
\end{figure*}

A key limitation in using these methods for a clinical population is the lack of large training datasets that are required to train robust and accurate models.  
While healthy participants can perform activities for long periods, stroke survivors fatigue rapidly, limiting the amount of data available per individual, and making data collection more costly and labor-intensive \cite{louie_exoskeleton_2020}.
Furthermore, post-stroke gait is highly heterogeneous \cite{mulroy_use_2003}, necessitating individualized models or large amounts of data to develop effective and generalizable data-driven estimations. 
To address the scarcity of clinical walking data, methods such as transfer learning \cite{pan_survey_2010} can leverage larger healthy walking datasets and supplement limited clinical datasets.
This could improve overall model performance without requiring extensive data collection, minimizing the data generation requirements for a population. 
Transfer learning may enable accurate torque estimates for individuals post-stroke, allowing for adaptive assistance that captures the variability in their gait. 

In this work, we leveraged transfer learning and multi-task learning to develop an approach for a machine-learning-based controller that delivers real-time adaptive dorsiflexion and plantarflexion exoskeleton assistance based on ankle torque estimates for post-stroke users. Specifically, we estimated ankle torque with a multi-task Temporal Convolutional Network (TCN), trained to simultaneously estimate ankle torque and stance phase. To test our data-driven method, we integrated it into a previously developed unilateral ankle exoskeleton \cite{cooper_design_2024}. To ensure a mobile system suitable for future community use, we rely on highly-portable body-mounted inertial measurement units (IMU) as inputs. 
In summary, in this work we:
\begin{itemize}
    \item Devised an optimized transfer and multi-task learning procedure for a TCN-based network architecture that yields feasible ankle torque estimates from IMU data despite the heterogeneous nature of post-stroke gait.
    \item Implemented a prototype for a portable system to assist post-stroke walking by augmenting an existing ankle robot with the capability to perform real-time torque estimates and use them for control. 
    \item Tested and evaluated our prototype on one stroke survivor walking on a treadmill.
\end{itemize}

\section{METHODOLOGY}

The estimation process was implemented as a supervised learning task (\Cref{fig:methods_overview}.A), in which ankle torque was predicted from a set of three IMUs and an ankle exoskeleton encoder. The ground truth kinematics and kinetics were computed from optical motion capture and ground reaction forces from an instrumented treadmill. To generate the real-time ankle torque estimates for control, we integrated a \mbox{Raspberry Pi 5} microcomputer (Raspberry Pi Foundation, Cambridge, UK) with an existing ankle exoskeleton.

\subsection{Augmenting an Ankle Exoskeleton for Real-Time Estimates}
\label{subsec:exo}

The exoskeleton used for this study (\Cref{fig:methods_overview}.B) is a previously developed portable, rigid ankle exoskeleton designed to assist post-stroke walking outside the laboratory \cite{cooper_design_2024}. This system features a distal actuation mechanism providing bidirectional assistance in both plantar and dorsiflexion. The waist-mounted pack houses a custom microcontroller unit (MCU), a lithium-ion battery, and an emergency stop system, with an adjustable belt and shoulder strap for user adaptability.

We placed three IMU sensors on the foot, shank, and thigh of the user’s paretic leg, and the exoskeleton's encoder was in line with the ankle joint. To enable real-time control commands, we incorporated a microcomputer into the system to take in the measurements from the IMUs and encoder and output an ankle torque estimate to the controller. The MCU of the exoskeleton then received the torque estimates from the microcomputer via CAN Bus at 100\;Hz. The low-level controller operated in a closed loop (1000\;Hz) with torque measurements from two load cells within the ankle exoskeleton as described in \cite{cooper_design_2024}. The controller fully relied on the estimated torque for plantarflexion and dorsiflexion commands. 

\begin{figure*}[thpb]
      \centering
      \includegraphics[width=0.95\linewidth]{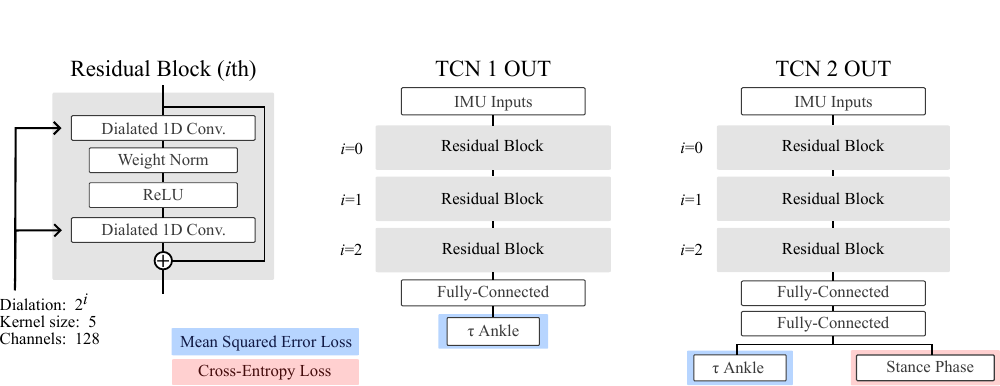}
      \caption{Our TCN architectures were composed of repeating Residual Blocks with increasing dilation factors. This design was closely aligned with the one proposed by \cite{molinaro_subject-independent_2022}. We evaluated a single-output version (TCN1OUT) and a dual-output version (TCN2OUT), which estimated both ankle torque and stance phase.
}
      \label{fig:tcn_architectures}
\end{figure*}

\subsection{Devising an Optimized Neural Network}

We developed an optimized neural network architecture to enable reliable, real-time ankle torque estimates using a microcomputer worn on the body. The authors of \cite{moghadam_comparison_2023} compared various machine learning approaches for predicting lower-limb joint kinematics and kinetics from IMU and electromyography (EMG) data and found that convolutional neural networks (CNN) and random forest (RF) machine learning approaches achieved the best results. Also, \cite{molinaro_subject-independent_2022} showed that temporal convolutional networks (TCN), deep learning models designed for sequence modeling \cite{bai_empirical_2018}, outperformed fully connected neural networks (FCNN) and long-short-term memory networks (LSTM) on hip torque estimation tasks using three IMUs. \cite{molinaro_estimating_2024} and then \cite{molinaro_task-agnostic_2024} further evaluated their TCN architecture across a wide range of dynamic tasks, highlighting its versatility.

As outlined in \cite{molinaro_subject-independent_2022,bai_empirical_2018}, TCNs process sequential data with 1D convolutions instead of recurrent operations, enabling efficient, parallelizable computation. Given the advantages offered by TCNs and their demonstrated effectiveness in comparable estimation tasks to this work, we selected the TCN architecture as the basis for our model.

\subsubsection{TCN Architectures}

We evaluated two TCN architectures as depicted in \Cref{fig:tcn_architectures}, one single-output architecture (TCN1OUT) and one dual-output architecture (TCN2OUT). Both models consisted of repeating Residual Blocks comprising two dilated 1-D convolutional layers with increasing dilation factors. Starting from $i=0$, the $i$th residual block was set up with the dilation factor $d_i=2^i$. This design closely mirrors the model architecture proposed by \cite{molinaro_subject-independent_2022,bai_empirical_2018}. In our framework, the TCN models took in a sequence of foot, shank, and thigh IMU sensor measurements and the estimated ankle torque at time $t$. Following the procedure outlined in \cite{molinaro_subject-independent_2022}, we set the input sequence length for both architectures according to the receptive field $h$ of the TCN, which was calculated as 
$$
h= 1+\sum^{l-1}_{i=0} 2(k-1)d_i,
$$
where $l$ is the number of residual blocks, $k$ is the kernel size, and $d$ is the dilation factor. An empirical grid search on the Unimpaired Dataset (\Cref{subsec:data}) yielded the hyperparameters $l=3$ residual blocks, with 128 channels and a kernel size $k=5$ for their convolutional layers. This resulted in an input sequence length $h=57$ for both network architectures.

\subsubsection{Multi-Task Learning}
In the dual-output TCN2OUT model, we hypothesized that multi-task learning would enhance ankle torque estimation by simultaneously training the network to classify gait phases (stance vs. swing). This approach is in principle justified, as the stance phase is a fundamental aspect of gait mechanics and directly influences ankle torques. By jointly optimizing for ankle moment (regression) and stance phase (classification), we aimed to guide the network towards a more physiologically meaningful representation of gait dynamics. As depicted on the right in \Cref{fig:tcn_architectures}, TCN2OUT featured two output heads. The first, identical to TCN1OUT, consisted of a single feature trained via optimizing Mean Squared Error (MSE) loss to predict ankle torque at time $t$. The second head comprised two features, encoding stance and swing phases using a one-hot representation. A softmax activation was applied to this output \cite{goodfellow_deep_2016}, where a higher value in the first neuron indicated stance phase, and a higher value in the second neuron indicated swing phase. This classification head was trained via optimizing Cross-Entropy Loss. 

When the ankle torque regression task and the stance phase classification task were trained jointly, the classification task often yielded a significantly larger loss, thereby dominating the optimization process. This imbalance, broadly classified as negative transfer, hindered learning and could cause less accurate ankle torques estimations for TCN2OUT compared to the single-task TCN1OUT. To mitigate negative transfer, we implemented exponentially moving average (EMA)-based loss balancing, following the method proposed in \cite{lakkapragada_mitigating_2023}. We define the individual EMA loss at training step $t$ for each output layer $k\in\lbrace1,2\rbrace$ as $\widetilde{L}_k(t)$, where $L_1(t)$ is the MSE Loss of the ankle torque regression, and $L_2(t)$ is the Cross-Entropy Loss of the stance phase classification:
$$
\widetilde{L}_k(t) = \alpha \cdot L_k(t) + (1-\alpha) \cdot \widetilde{L}_k(t-1).
$$
We set the smoothing factor $\alpha=0.9$. The balanced multi-task learning loss was then estimated as the sum of the individual losses normalized by their EMA losses: 
$$
\mathrm{Loss}(t) = \sum^2_{k=1}\frac{L_k(t)}{\widetilde{L}_k(t)}.
$$
We compared TCN1OUT and TCN2OUT ankle torque estimates through root mean squared error (RMSE), mean squared error (MAE), and coefficient of determination ($\mathrm{R}^2$). Further, we specifically compared differences in RMSEs during the stance and swing phase of both architectures and stance phase classification accuracy of the TCN2OUT. 

\subsubsection{Pretraining and Fine-Tuning}
\label{subsec:pretrain}

We hypothesized that including training data from healthy populations would improve ankle torque estimates for post-stroke, by introducing the model to the biomechanics of the leg and how they relate to attached IMU sensors on larger data variations. For data collection, unimpaired participants could walk more, at a wider range of speeds, and were comfortable with speed transitions. The relative ease to collect unimpaired data resulted in data imbalance, in this work, outsizing the impaired data by a 10 to 1 ratio. Our datasets will be detailed in the following \Cref{subsec:data}. To mitigate data imbalance, we investigated a finetuning approach, where we evaluated four model configurations. The first two were TCN1OUT and TCN2OUT trained on post-stroke data only. The remaining two were TCN1OUT and TCN2OUT pretrained on unimpaired data and then fine-tuned by continuing to train all weights and biases on the smaller set of post-stroke data. 

For the model evaluation, we performed leave-one-out cross-validation. This approach involved training separate models, each time leaving out the data of one participant as the test set, while the remaining data was split into training and validation sets with a 90\%-10\% ratio. For pretraining on unimpaired data, we performed a random train and validation set split at a 90\% to 10\% ratio of all unimpaired data. 

For all training procedures, we utilized the AdamW optimizer with a learning rate of 0.0001, and a batch size of 32. Pre-training involved training the model for 80 epochs on the unimpaired data first. For model training or fine-tuning on post-stroke data, we split the data into training and test sets by participant. We trained for 80 epochs with the same optimizer settings and batch sizes used in the model trained on the unimpaired dataset.

\subsection{Data}
\label{subsec:data}

\begin{figure}[t]
      \centering
      \includegraphics[]{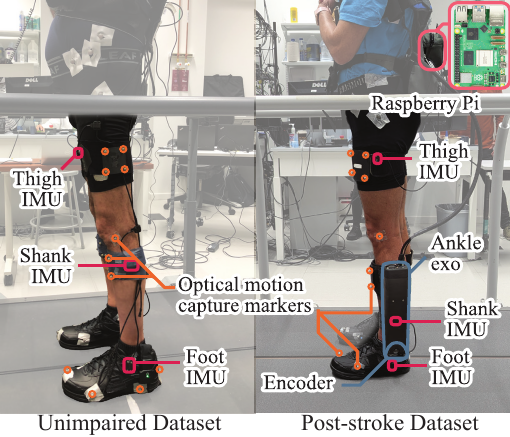}
      \caption{We collected a pretraining dataset on a healthy population (Unimpaired Dataset) and a dataset collected from individuals in the chronic phase of stroke (Post-stroke Dataset). Left shows the unimpaired left-leg sensorization. Right shows the ankle exoskeleton and post-stroke participant sensorization.
}
      \label{fig:data_collection}
\end{figure}

A main challenge emerged from natural variability in the sensor data and limited training data for post-stroke gait. Compared to the large datasets \cite{molinaro_task-agnostic_2024} collected on healthy participants, post-stroke data collection is more costly, in part due to the high fatigability of stroke survivors. Therefore, to accumulate sufficient data for training our TCN architectures in this work, we collected two datasets; one larger pretraining dataset from a healthy population, and a smaller fine-tuning dataset from a population in the chronic stage of stroke recovery. All data collection was approved by the Harvard Longwood Medical Area Institutional Review Board under the ID \#IRB16-1845 and all individuals provided medical clearance and written informed consent. 

Both datasets contained the following data collection modalities. An optical motion capture system (Qualisys, Gothenburg, Sweden; 200\;Hz) was used to measure kinematics. The collected data included orientation angles, and angular velocities for the foot, shank, and thigh of the paretic limb. Additionally, Ground Reaction Forces (GRFs) were measured from an instrumented split-belt treadmill (Bertec, Columbus, OH, USA; 2000\;Hz). Both the optical motion capture and GRF data served to estimate ground truth ankle moment through inverse dynamics using the Visual3D Software (HAS-Motion, Kingston, Ontario, Canada) \footnote{\url{https://www.has-motion.ca}}. Further, participants were equipped with Inertial Measurement Units (MTi-3, Movella Inc., Nevada, USA) at 100\;Hz on their foot, shank, and thigh segments. From all IMUs, we recorded intrinsic global orientation estimates in Euler angles ($\mathbb{R}^3$), triaxial acceleration ($\mathbb{R}^3$), and triaxial gyroscope data ($\mathbb{R}^3$). As depicted in \Cref{fig:data_collection}, IMU sensor placement and the data collection protocol varied across the unimpaired and post-stroke datasets.

\subsubsection{Unimpaired Dataset}

The Unimpaired Dataset was collected from individuals without neuromotor gait impairment (6 participants, 4\;M\;/\;2\;F, age $29\pm3.9$\;yrs). We sensorized their left leg, as all participants post-stroke were left-paretic. As shown in \Cref{fig:data_collection} on the left, all the IMUs were laterally positioned on the participants’ left leg, aligned with the sagittal plane. Participants walked on the instrumented treadmill for 6 trials of 4 minutes each. Each trial featured randomized speed conditions ranging from 0.6\;m/s to 1.4\;m/s. For the first 2 minutes, the treadmill maintained a constant speed of 1\;m/s before ramping up to either 1.2\;m/s or 1.4\;m/s, or down to 0.8\;m/s or 0.6\;m/s for the final 2\;minutes. Half of the participants started with acceleration, while the others began with deceleration. The speed variations were implemented to capture a range of walking patterns and enhance dataset diversity. On average, $2118\pm442$ strides were collected per participant, resulting in a total of $12710$ strides. 

\subsubsection{Post-stroke Dataset}

The Post-stroke Dataset was collected from 4 left-paretic stroke survivors (4\;M, age $55\pm5.5$\;yrs). Participants wore the ankle robot on their paretic limb. Each participant completed 6 trials on the instrumented treadmill, comprising of two 2\;min trials for each a self-selected comfortable walking speed (CWS: $0.65\pm0.13$\;m/s), a slow walking speed (SWS: $0.5\pm0.08$\;m/s), and a fast walking speed (FWS: $0.8\pm0.18$\;m/s). Participants post-stroke demonstrated high variability in their self-selected CWS, which ranged from 0.5 to 0.8\;m/s. On average, each participant performed $334\pm79$ strides, which resulted in a total of $1334$ strides.

Post-stroke sensorization is depicted in \Cref{fig:data_collection} on the right. One IMU was embedded in the ankle exoskeleton on the lateral side of the shank, and the robot provided an ankle angle estimate from the integrated encoder. The remaining IMUs were placed on the lateral side of the paretic foot and on the lateral side of the thigh. The robot was active and delivered plantarflexion and dorsiflexion assistance during these trials. We utilized inverse dynamics of Visual3D to compute the net ankle torque from optical motion capture data and ground reaction forces, which is the sum of the torque contributions from the biological torque and as well as applied torque by the robot. 

\subsubsection{Data Merging and Post-Processing}

For both datasets, all data sources were time synchronized: IMUs at 100\;Hz, GRFs at 2000\;Hz, optical motion capture at 200\;Hz, and the ankle robot encoder at 100\;Hz. Our preprocessing pipeline aligned IMU data with kinetic measurements (ankle moment, vertical GRF) and marker trajectories from the optical motion capture system. All synchronized signals were then resampled to\;100 Hz to match the IMU sampling rate.
Since IMU sensor placement differed between Unimpaired and Post-stroke Datasets, we further post-processed the unimpaired IMU data by applying a rotational transformation to align it with the reference frames in the Post-stroke Dataset. To approximate encoder measurements of the ankle exoskeleton, we used the sagittal angle from the unimpaired foot IMU as a proxy. We then filtered ground-truth ankle torques with a fourth-order, 10\;Hz low-pass filter to obtain smoother ankle moment estimates. This reduced high-frequency noise while preserving the essential gait characteristics, thus improving the stability of our neural network predictions. For model training, all IMU sensor measurements were standardized using the means and standard deviations of the training data.

\subsection{Exoskeleton Application}

Alongside the outlined evaluation on collected datasets, a key contribution of this paper is demonstrating an end-to-end prototype for real-time adaptive torque control of an ankle exoskeleton. For this demonstration, under the same IRB protocol (\#IRB16-1845) with medical clearance and written informed consent, we invited a participant post stroke to the laboratory and asked them to complete two walking trials at their self-selected treadmill comfortable walking speed while using our prototype. For this application, we implemented the following design choices and safeguards for the final control commands to the robot. 

\subsubsection{Input Channels and Feature Importance}

To maintain a reliable 100\;Hz throughput on our microcomputer without compromising performance, we limited the amount of transmitted data. Instead of streaming all available sensor data, we performed a preliminary feature importance analysis to lower computational costs and minimizing required IMU channels. Our IMUs streamed at a total of 27\;channels, 9\;signals per IMU, which were triaxial angles, accelerations, and angular velocities. We aimed to reduce the input feature size to 16, such that, considering the previously determined perceptive field of 57 observations, the input to TCN1OUT and TCN2OUT would be $\mathbb{R}^{16\times57}$.

Using our larger Unimpaired Dataset, we trained six baseline TCN1OUT networks to evaluate ankle torque estimations from all available IMU data in leave-one-out cross validation. We assessed feature importance by measuring the effect on prediction accuracy when randomizing either all IMU angles, all accelerations, or all gyroscope readings in the test data. We then removed the channels that had the least influence on prediction accuracy from our training data in successive model optimizations for devising the best-performing architecture. The resulting streamed channels were the angles and angular velocities of thigh and shank IMUs, the angular velocities of the foot IMU, and its sagittal angle in case of the Unimpaired Dataset, or the ankle exoskeleton encoder in case of the Impaired Dataset.

\subsubsection{Testing The Prototype}
Given the vulnerability of individuals poststroke and the black-box nature of neural networks, we took safety precautions during the exoskeleton trial. To avoid exposing the user to artifacts from noisy estimates, the ankle torque estimate from the microcomputer was low-pass filtered by a second-order Infinite Impulse Response Butterworth filter set at a 7 Hz cutoff frequency. Given our estimates were produced at 100 Hz, this resulted in a phase delay of about 35.7 ms, which is comparable to related works \cite{molinaro_task-agnostic_2024}. The proportional gain was set to 20\% of the participant’s mass. Further, the participant wore a ceiling-mounted harness for fall prevention. The participant completed two walking trials at their self-selected treadmill comfortable walking speed. We evaluated the trial by assessing the quality of real-time torque predictions using MAE, RMSE, and $\mathrm{R}^2$ scores. Additionally, we compared the timing and magnitude of estimated torque peaks and troughs against ground-truth ankle torque measurements to assess the system's accuracy and responsiveness.

\section{RESULTS AND DISCUSSION}

\subsection{Feature Importance}
\begin{table}[h]
\caption{Effects when randomizing sensors (highlighted lowest)
}
\label{tab:feature_importance}
\begin{center}
\begin{threeparttable}
\begin{tabular}{p{0.7cm} c c c c}
\toprule
     & Baseline &Euler Angles& Accel. & Gyro. \\
\midrule
MAE& $0.11\pm0.04$&$+0.13\pm0.04$ & \cellcolor{black!15}$+0.03\pm0.02$&$+0.07\pm0.03$ \\
RMSE& $0.16\pm0.05$&$+0.19\pm0.06$ & \cellcolor{black!15}$+0.04\pm0.03$&$+0.11\pm0.04$ \\ 
$\mathrm{R}^2$& $0.81\pm0.16$&$-0.57\pm0.24$ & \cellcolor{black!15}$-0.09\pm0.05$&$-0.25\pm0.34$ \\ 
\bottomrule
\end{tabular}
\begin{tablenotes}\footnotesize
\item[*] Units for error measures are Nm/kg
\end{tablenotes}
\end{threeparttable}
\end{center}
\end{table}

We conducted a feature importance analysis to reduce the number of streamed channels. The results are summarized in \Cref{tab:feature_importance}. Randomizing the acceleration measurements increased RMSE by $+0.04\pm0.03$\;Nm/kg, the MAE by $+0.03\pm0.02$\;Nm/kg, and decreased the $\mathrm{R}^2$ score by $-0.09\pm0.05$ on average. Despite accelerations being integral to inverse dynamics, their omission affected prediction accuracy the least, compared to orientation and gyroscope data. We excluded the accelerations from our subsequent model training to ensure the targeted 100\;Hz throughput on the microcomputer. We attribute the relatively low impact of omitting accelerations to the strictly cyclic nature and low variability of walking at a constant speed on a treadmill. Accelerations remain of critical importance for data-driven control when extending to diverse tasks, which is supported by their theoretical relevance and incorporation by related work \cite{molinaro_task-agnostic_2024, molinaro_estimating_2024}. Nevertheless, research has explored gait event detection from solely gyroscope data \cite{tong_practical_1999,greene_adaptive_2010}, and our results utilizing angular information may warrant further exploration for sensor minimization in strictly cyclic tasks.

\subsection{Optimal Model}

\begin{figure}[t]
      \centering
      \includegraphics[width=\linewidth]{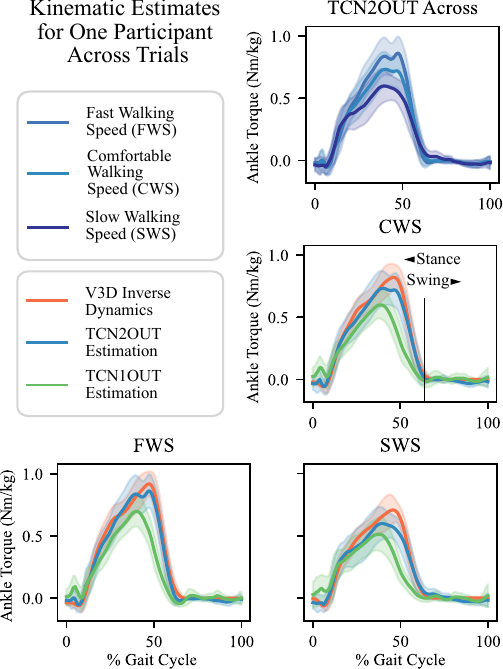}
      \caption{The top-most plot shows the TCN2OUT estimates across all three walking conditions, demonstrating the adaptiveness of the model to various speeds. The three lower plots illustrate ankle torque predictions of TCN2OUT and TCN1OUT across the fast (FWS), comfortable (CWS) and slow walking conditions (SWS) of an example post-stroke participant. Orange lines are the ground truth obtained from optical motion capture and force plates. 
}
      \label{fig:one_p_results}
\end{figure}

\begin{table}
\caption{Model comparison (highlighted best average)
}
\label{tab:model_comp}
\begin{center}
\begin{threeparttable}
\begin{tabular}{r c c c c}
\toprule
 & \multicolumn{2}{c}{TCN2OUT} & \multicolumn{2}{c}{TCN1OUT}\\
 \cmidrule(l{3pt}r{3pt}){2-3} \cmidrule(l{3pt}r{3pt}){4-5}
 & Fine-tuned & \begin{tabular}{@{}c@{}}Post-stroke\\only\end{tabular} & Fine-tuned & \begin{tabular}{@{}c@{}}Post-stroke\\only\end{tabular} \\
\midrule
MAE & \cellcolor{black!15}$0.11\pm0.04$ & $0.13\pm0.05$ & $0.13\pm0.04$ & $0.14\pm0.06$ \\
RMSE & \cellcolor{black!15}$0.16\pm0.04$ & $0.18\pm0.05$ & $0.18\pm0.04$ & $0.18\pm0.06$ \\
$\mathrm{R}^2$ & \cellcolor{black!15}$0.74\pm0.13$ & $0.65\pm0.16$ & $0.66\pm0.10$ & $ 0.65\pm0.22$ \\
St. RMSE & \cellcolor{black!15}$0.19 \pm 0.05$ & $0.22 \pm 0.05$ & $0.22 \pm 0.04$ & $0.21 \pm 0.06$\\
Sw. RMSE & \cellcolor{black!15}$0.05 \pm 0.03$ & $0.08 \pm 0.05$ & $0.06 \pm 0.03$ & $0.10 \pm 0.09$\\
Phase Acc. & \cellcolor{black!15}$0.90\pm0.04$ & $0.87 \pm 0.07$ & - & - \\
\bottomrule
\end{tabular}
\begin{tablenotes}\footnotesize
\item[*] Units for error measures are Nm/kg
\end{tablenotes}
\end{threeparttable}
\end{center}
\end{table}

A comparison of our four model types is presented in \Cref{tab:model_comp}. Overall, the fine-tuned TCN2OUT showed the lowest prediction errors and highest $\mathrm{R}^2$ scores in our leave-one-out cross-validation on all four post-stroke participants. \Cref{fig:one_p_results} showcases adaptability of both TCN1OUT and TCN2OUT to changes in walking speed for one participant in our test dataset.

The results indicate that TCN-based ankle moment estimation is promising in the context of existing literature. \cite{simonetti_wearable_2024} estimated post-stroke ankle dorsi-plantarflexion torque using an EMG-driven musculoskeletal model combined with data from five IMUs. They reported leave-one-out accuracies of $\mathrm{R}^2$ $0.67\pm0.22$ and MAE $0.32\pm0.17$\;Nm/kg at comfortable walking speeds (CWS) and $\mathrm{R}^2$ $0.63\pm0.20$ and MAE $0.39\pm0.15$\;Nm/kg at fast walking speeds (FWS). Our TCN models achieved comparable or improved performance while relying on only three IMUs placed on the paretic limb. However, this comparison serves only as an indication, as we validated our model on a different dataset. Similarly, \cite{moghadam_comparison_2023} estimated ankle torques in healthy individuals walking overground using EMGs and seven IMUs. They reported leave-one-out errors of MAE $0.158\pm0.087$\;Nm/kg ($\mathrm{R}^2$=74) when using a CNN and MAE $0.117\pm0.083$\;Nm/kg ($\mathrm{R}^2$=84) with a Random Forest model. Although their results were obtained on a different population and sensor setup, our work suggests that data-driven methods using fewer sensors can still provide accurate torque estimates for post-stroke populations.

Fine-tuning a model pretrained on the Unimpaired Dataset had an observable positive effect on leave-one-out model performance for TCN2OUT. Without pretraining, TCN2OUT achieved $\mathrm{R}^2$ $0.65\pm0.16$, whereas pretraining increased this to $\mathrm{R}^2$ $0.74\pm0.13$, which is the highest among all compared models. This suggests that pretraining helped TCN2OUT optimize both ankle moment estimation and stance phase classification, supporting our hypothesis that learning both tasks may have encouraged the model to develop a more structured, physiologically meaningful representation of gait dynamics.

\subsection{Exoskeleton Application}

\begin{figure}[t]
      \centering
      \includegraphics[width=1\linewidth]{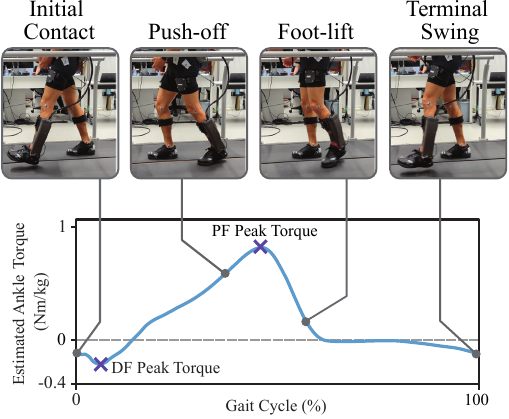}
      \caption{Representative snapshots of the estimated ankle torque over a single stride during the exoskeleton walking trial at four key points in the gait cycle: initial contact, push-off, foot-lift, and terminal swing. The peak dorsiflexion (DF) torque during weight acceptance and peak plantarflexion (PF) torque during terminal stance are highlighted on the torque curve.
}
      \label{fig:real_robot}
\end{figure}

\begin{figure}[t]
      \centering
      \includegraphics[width=\linewidth]{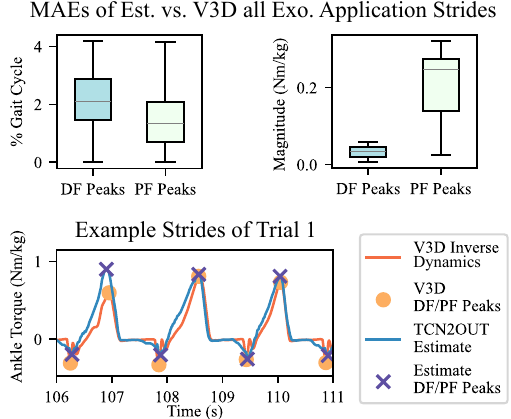}
      \caption{\textbf{Top}: Summarized errors for magnitudes and \% gait cycle difference for estimated dorsiflexion (DF) and plantarflexion (PT) Peaks for all strides of our exoskeleton trial. \textbf{Bottom}: Example strides of the first exoskeleton trial. The orange line indicates the ground truth ankle torque estimated through inverse dynamics from optical motion capture and force plate data. The blue line represents the TCN2OUT estimates utilized as the controller inputs in this trial. 
}
      \label{fig:peaks_plot}
\end{figure}

Testing our prototype, we collected 159 strides at comfortable  speed from two walking bouts. We utilized the TCN2OUT for ankle torque estimates, which achieved an overall RMSE of \mbox{0.14\;Nm/kg}, MAE 0.1\;Nm/kg, and $\mathrm{R}^2$ of 0.63. These results are comparable to results on our Post-stroke Dataset (\Cref{tab:model_comp}). We evaluated observed dorsiflexion (DF) and plantarflexion (PF) peaks timing in \mbox{\% gait cycle} and magnitude. These measurements are of high importance for exoskeleton controllers as they decide when and how much assistance is applied. As depicted in \Cref{fig:real_robot}, these peaks occur around initial contact and push-off in the gait cycle. The results are summarized in \Cref{fig:peaks_plot}. The PF peaks exhibited a notable error of $0.24\pm0.09$\;Nm/kg. The timing error of both peaks however was promisingly low with $1.94\pm0.97$\;\mbox{\% gait cycle} deviation for DF peaks and $1.56\pm0.95$\;\mbox{\% gait cycle} for PF peaks. Given the low error in timing, the higher magnitude error was not a major concern for our exosuit trial, as it can be adjusted through proportional gain by the controller. Considering the relatively small population of four post-stroke participants in our training data, we expect that training future estimation models on larger populations could improve these results further. 

Testing data-driven assistive exoskeletons on post-stroke populations is challenging due to the necessary safety requirements and limited data available. This challenge also affected data collection for this work. Evaluation of our presented prototype included a limited set of post-stroke participants and treadmill walking. Further, our ankle exoskeleton application tested only one speed condition. Future research should approach the challenging task of gathering and merging larger post-stroke datasets, to validate across larger populations, overground walking, and a broader range of activities. Furthermore, direct command of a scaled estimated torque is not optimal for participants with ankle weakness. Future work will explore methods to optimize assistance based on estimated torques.

Our evaluation involved ankle torque estimates for the entire stride, including swing-phase. Although swing phase predictions carry limited significance due to the relatively low ankle torque the paretic limb produces during swing, they remain critical for ankle dorsiflexion and toe clearance. As observed in \Cref{tab:model_comp}, TCN2OUT achieved lower \mbox{RMSEs} for swing phase predictions ($0.05\pm0.034$\;Nm/kg for fine-tuned models, \mbox{$0.08\pm0.057$\;Nm/kg} for post-stroke-only models). However, the ground-truth data used for training was not fully representative of true swing-phase ankle torque due to limitations in modeling the inertial properties of the body segments used to compute inverse dynamics. Future work could address this by incorporating the stance phase predictions of a dual-output model like TCN2OUT to affect how the controller utilizes the ankle torque estimate.


\section{CONCLUSIONS}

We see the demonstration of this work as a promising step toward translating data-driven advancements in adaptive exoskeleton control from healthy individuals to post-stroke populations, paving the way for more accessible and effective mobility assistance. Results on the data of four people post stroke demonstrate the validity for assistance, which is adaptive to individual kinematic and speed conditions. Our model was optimized for real-time performance and we see further validation in the application of our prototype in an ankle exoskeleton trial, where we used ankle torque estimates as inputs for a high-level controller. Evaluation was limited to treadmill walking data of four people post stroke and one prototype exoskeleton application. Future work will focus on evaluating larger populations, and broader activity ranges.


\section*{ACKNOWLEDGMENTS}
The authors thank Krithika Swaminathan, Teresa Baker, Tara Kimiavi, and our participants for their time and contributions to this study. This work was supported by funding from the Move Lab at the Harvard School of Engineering and Applied Sciences, a Massachusetts Technology Collaborative, Collaborative Research and Development Matching Grant and the Raj Bhattacharyya and Samantha Heller Assistive Technology Initiative Fund.

\bibliographystyle{IEEEtran}
\scriptsize{
\bibliography{real_time_ankle_moment_exo}
}

\end{document}